\documentclass[letterpaper, 10 pt, conference]{ieeeconf}  

\IEEEoverridecommandlockouts                              

\usepackage{amsmath} 
\usepackage{amssymb}  

\usepackage[final]{graphicx}  
\usepackage{bm}
\usepackage{cite}
\usepackage[tracking]{microtype}
\usepackage{csquotes}
\usepackage[draft,inline,nomargin,index]{fixme}
\usepackage{mathtools}
\usepackage[super]{nth}
\usepackage{makecell} 
\usepackage[shortcuts, acronym, toc, nonumberlist=false]{glossaries}
\usepackage{multirow}
\usepackage{dsfont}
\usepackage[percent]{overpic}
\usepackage{tikz}

\newcommand{\ie}{i.\,e.,\ }
\newcommand{\eg}{e.\,g.,\ }
\newcommand{\wrt}{w.\,r.\,t.\ }

\usepackage{booktabs}
\usepackage{boldline}
\usepackage{siunitx}
\DeclareSIUnit{\rad}{rad}

\newacronym
 {rmse}{RMSE}{Root Mean Square Error}
\newacronym
 {fde}{FDE}{Final Displacement Error}
\newacronym
 {sl}{SL}{Supervised Learning}
\newacronym
 {rl}{RL}{Reinforcement Learning}
\newacronym
 {il}{IL}{Imitation Learning}
\newacronym
 {bc}{BC}{Behavior Cloning}
\newacronym
 {cv}{CV}{Constant Velocity}
\newacronym
 {pomdp}{POMDP}{Partially-Observable Markov Decision Process}
\newacronym 
 {ppo}{PPO}{Proximal Policy Optimization}
\newacronym 
 {gae}{GAE}{Generalized Advantage Estimation}
\newacronym
 {airl}{AIRL}{Adversarial Inverse Reinforcement Learning}
\newacronym
 {gan}{GAN}{Generative Adversarial Network}
\newacronym
 {mlp}{MLP}{multi-layer perceptron}
\newacronym
 {hdmap}{HD Map}{High-Definition Map}
\newacronym
 {sft}{SFT}{Symmetric Fusion Transformer}
\newacronym
 {rpe}{RPE}{Relative Positional Embedding}
\newacronym
 {mp}{MP}{Message Passing}
\newacronym
 {mhca}{MHCA}{Multi-Head Cross Attention}
\newacronym
 {isps}{ISPS}{Inference Steps per Second}
\newacronym
 {vru}{VRU}{Vulnerable Road User}
\DeclarePairedDelimiterX{\infdivx}[2]{(}{)}{%
  #1\;\delimsize\|\;#2%
}

\usepackage[draft,inline,nomargin,index]{fixme}
\fxsetup{theme=color, mode=multiuser}
\FXRegisterAuthor{f}{fabian}{\color{red}Fa}

\newif\iffde
\fdetrue 

\title{\LARGE \bf
Toward Efficient and Robust Behavior Models for \\Multi-Agent Driving Simulation
}

\author{Fabian Konstantinidis,
        Moritz Sackmann,
        Ulrich Hofmann,
        Christoph Stiller
\thanks{Fabian Konstantinidis is with CARIAD SE, Germany, and also with Karlsruhe Institute of Technology, Germany.  
Moritz Sackmann is with CARIAD SE, Germany.
Christoph Stiller is with Karlsruhe Institute of Technology, Germany.
This work is a result of the joint research project STADT:up (19A22006E). The project is supported by the German Federal Ministry for Economic Affairs and Climate Action (BMWK), based on a decision of the German Bundestag. The author is solely responsible for the content of this publication.}%
}

\begin{document}

\maketitle
\thispagestyle{empty}
\pagestyle{empty}

\begin{abstract}
Scalable multi-agent driving simulation requires behavior models that are both realistic and computationally efficient. We address this by optimizing the behavior model that controls individual traffic participants. To improve efficiency, we adopt an instance-centric scene representation, where each traffic participant and map element is modeled in its own local coordinate frame. This design enables efficient, viewpoint-invariant scene encoding and allows static map tokens to be reused across simulation steps. To model interactions, we employ a query-centric symmetric context encoder with relative positional encodings between local frames. We use Adversarial Inverse Reinforcement Learning to learn the behavior model and propose an adaptive reward transformation that automatically balances robustness and realism during training. Experiments demonstrate that our approach scales efficiently with the number of tokens, significantly reducing training and inference times, while outperforming several agent-centric baselines in terms of positional accuracy and robustness.
\end{abstract}

\section{INTRODUCTION}

Given the high cost and potential risks of deploying automated vehicles in real-world environments, simulation is indispensable for the research and development of advanced driver assistant systems. Key applications include generating diverse scenarios that may be difficult or dangerous to recreate in the real world \cite{suo2023mixsim, chitta2024scenario_generation2, amini2020RAL_RL_edge_case}, predicting human driving behavior \cite{zhang2023trafficbots, bhattacharyya2022modeling, Konstantinidis2024IV}, and training or evaluating behavior models \cite{apple2025self-play, cornelisse2025self-play}.

A key requirement for multi-agent driving simulation is exhibiting realistic behavior to reduce the sim-to-real gap. For that, behavior models map observations to actions. They are typically trained via \gls{il}, \eg \cite{bansal2018chauffeurnet, lu2023imitation}, which learns directly from real data but yields less robust policies, or via \gls{rl}, \eg \cite{apple2025self-play, chen2019model, konstantinidis2023ITSC, arief2024RAL_IS_meta_rl}, which improves robustness through exploration in simulation but depends on a reward function, which is unknown for real-world drivers.

Instead, we employ \gls{airl} \cite{fu2018airl} to reconstruct a reward signal from real traffic data and simultaneously learn a behavior model that maximizes this reward signal using standard \gls{rl}. Since the reconstructed reward signal can differ across experiments, it may unintentionally affect the robustness of the model. To address this, we introduce an adaptive reward transformation that balances robustness and realism during training.

\begin{figure}[t]
  \centering
  \includegraphics[width=0.38 \textwidth]{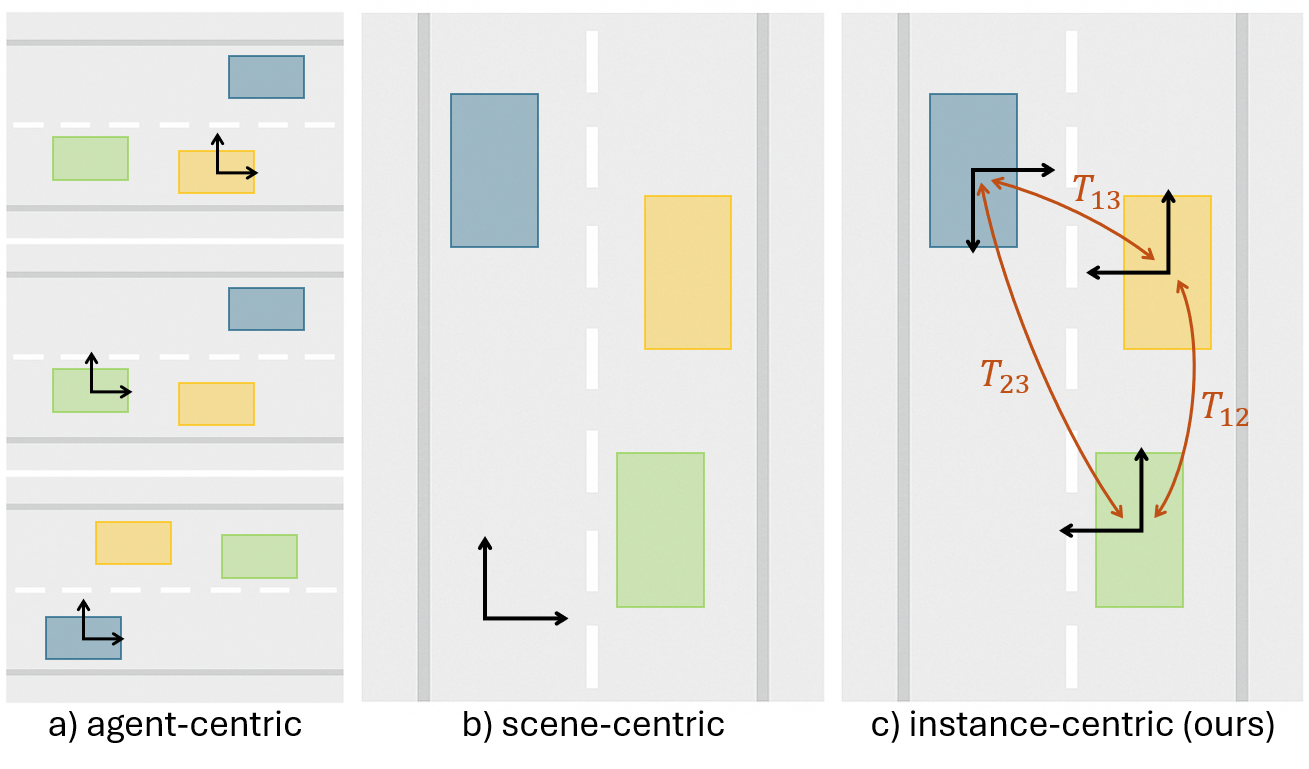}
  \vspace{-9pt}
  \caption{Illustration of different scene representations. Our instance-centric representation encodes each instance in its own coordinate frame, enabling shared feature extraction.}
  \label{fig:obs_representation}
  \vspace{-16pt}
\end{figure}

The second requirement is achieving high execution speed to enable efficient scaling. While prior works, \eg \cite{gulino2023waymax, kazemkhani2024gpudrive, scibior2021torchdrivesim, apple2025self-play}, have focused on accelerating the simulation backbone, our focus is on optimizing the executed behavior model.

Commonly, agent-centric observations (see Figure \ref{fig:obs_representation}a) are used, \eg\cite{Konstantinidis2024IV, kazemkhani2024gpudrive, apple2025self-play, cornelisse2025self-play}, where nearby instances, such as agents and static map elements, are represented from each agent's perspective. However, generating and encoding these local observations for every agent at each simulation step imposes a significant computational burden. Alternatively, scene-centric observations (see Figure \ref{fig:obs_representation}b) represent all instances in a global coordinate frame, \eg \cite{suo2021trafficsim, ngiam2021scene}. While this offers efficient encoding, it comes at the potential sacrifice of pose-invariance, requiring the model to implicitly learn variations in agent positions and orientations.

Instead, to improve the efficiency of the behavior model, we employ instance-centric observations (see Figure \ref{fig:obs_representation}c), where each instance, \ie an individual agent or map element, is represented in its own local coordinate frame. Features are encoded using shared context encoders, allowing the encoded instance tokens to be processed across multiple agents. This approach models relationships between tokens in a symmetric manner, independent of any global coordinate frame. As a result, static instances need to be encoded only once, and dynamic instances only once per simulation step, rather than from every agent's perspective at each step.

\emph{\textbf{Our Contributions}} are four-fold:
(1) Our main contribution is the introduction of an instance-centric scene representation for behavior modeling, which enables efficient, viewpoint-invariant encoding of the scene. Instance tokens are shared across all agents. Static instances need to be encoded only once, enabling the reuse of their tokens in subsequent simulation steps.
(2) We employ a query-centric symmetric context encoder with relative positional encodings between pairwise instance-centric frames, and evaluate its performance across different model design choices. 
(3) We propose an adaptive reward transformation that automatically balances robustness and realism during \gls{airl} training.
(4) We demonstrate that our approach yields realistic behavior models, outperforming a diverse set of baselines in robustness and scalability across two automated driving datasets.

\section{Related Work}
Achieving both realism and high execution speed is crucial for modern traffic simulations. While high-fidelity behavior modeling ensures the accuracy of the simulated traffic flow, the ability to execute such models efficiently remains equally critical, especially in large-scale or real-time applications.

\emph{\textbf{Realistic Behavior Simulation}}: 
Early approaches to simulating vehicle behavior include log-replay \cite{bergamini2021simnet, gulino2023waymax}, where trajectories recorded in the real world are simply replayed, rule-based models \cite{kesting2007general, kesting2010enhanced, gulino2023waymax}, as well as \gls{bc} \cite{lu2023imitation, bansal2018chauffeurnet}, an open-loop \gls{il} method, where a mapping from observations to actions is learned from a dataset using supervised learning. Due to its simplicity and the availability of large-scale datasets, the latter remains widely used. Each approach, however, has its limitations: Replayed trajectories are non-reactive, thus prohibiting interactions with a simulated ego vehicle. Rule-based models often fail to capture the complexity of real-world behaviors. \gls{bc} performs poorly in rare or unseen scenarios (\eg near-collisions or very aggressive merges). Even worse, when deploying such a model in a closed-loop simulation, small deviations from expert driving accumulate, leading the model into unfamiliar states - known as \emph{covariate shift} \cite{spencer2021feedback}.

For learning more robust behavior models, \gls{rl} is commonly employed \cite{chen2019model, cornelisse2025self-play, apple2025self-play, konstantinidis2023ITSC, arief2024RAL_IS_meta_rl}. 
Since realistically simulating traffic scenarios is inherently a multi-agent problem, self-play \gls{rl} has emerged as an effective solution. In \cite{konstantinidis2023ITSC}, all agents are controlled by a single shared policy, which is trained jointly through self-play. To enable applying the model to a diverse set of traffic scenarios, a flexible, graph-based input representation with vectorized map features is adopted. Similarly, in \cite{apple2025self-play} and \cite{cornelisse2025self-play}, self-play \gls{rl} has been applied on a large scale, demonstrating that robust and naturalistic behavior emerges from it. The resulting behavior model generalizes well, even in out-of-distribution scenarios. Notably, in \cite{apple2025self-play}, the behavior model is trained in randomly initialized traffic scenarios, driving more than \num{1.6} billion km during training. The resulting behavior model is zero-shot evaluated on three autonomous driving benchmarks, achieving state-of-the-art performance without ever seeing human data during training.

However, achieving realistic behavior with \gls{rl} requires a carefully tuned reward function that precisely guides the agents toward human-like driving. Recent works \cite{Konstantinidis2024IV, sun2024modelling, weaver2024RAL_residualAIL, bhattacharyya2022modeling} address this issue by reconstructing a reward signal from real-world data. In \cite{Konstantinidis2024IV}, \gls{airl} is used, where a discriminator is trained to distinguish real from simulated behavior, assigning higher scores to more realistic samples. As the goal is to drive as realistically as possible, the output of the discriminator is then used as a reward signal for \gls{rl} training. Similarly, we train our model using \gls{airl}, but introduce an adaptive reward transformation to balance robustness and realism.

\emph{\textbf{Enhanced inference speed}}:
Many existing works on multi-agent driving simulation focus on optimizing the simulation backbone, \ie the engine responsible for executing actions and producing next-step observations. This is commonly done by implementing the simulator in PyTorch \cite{scibior2021torchdrivesim, apple2025self-play}, JAX \cite{gulino2023waymax}, or C++ using CUDA \cite{kazemkhani2024gpudrive} to support in-graph compilation for hardware (GPU/TPU) acceleration. 
However, we argue that often, not the simulation engine, but the executed behavior model is the limiting factor. We therefore focus on optimizing the executed behavior model for multi-agent simulation.

All the discussed approaches to behavior modeling employ agent-centric observations, where each observation is expressed relative to the simulated agent's frame of reference. While straightforward, this becomes computationally expensive when simulating multiple agents, as the complexity scales linearly with the number of agents and quadratically with the number of pairwise interactions between agents.

Although coordinated scene-centric agent behavior has already been investigated for single-shot motion forecasting, \eg \cite{ngiam2021scene, su2022narrowing}, to the best of our knowledge, TrafficSim \cite{suo2021trafficsim} is the only work on learning a scene-centric multi-agent behavior model for closed-loop simulation. In \cite{suo2021trafficsim}, a global map is rasterized and encoded using a CNN. Then, local map features are extracted via Rotated Region of Interest Align and fused with the agent features. Lastly, a joint decoder model, realized as a message passing network, processes all agent features jointly, enabling coordination between them. This scene-centric representation effectively leads to more efficient simulations, as the static environment needs to be encoded only once, allowing the reuse of its encoded tokens in subsequent simulation steps. However, due to the global coordinate frame, this comes with the potential sacrifice of pose-invariance, often leading to performance degradation when the global coordinate frame changes.

To address this, recent single-shot motion prediction methods, \eg \cite{zhang2024simpl, zhou2023query, zhang2023hptr, shi2024mtr++}, adopt instance-centric representations, encoding each agent and map element in its own local frame. These pose-invariant features can be reused across all target agents, with relative positions enabling symmetric encoding from any agent’s perspective.

To the best of our knowledge, we are the first work to employ instance-centric representations for closed-loop traffic simulation, enabling us to accelerate the simulation by reusing encoded tokens of static map elements across simulation steps.

\section{Method}
For learning and evaluating behavior models, a simulation framework is required. In our in-house simulation framework, each vehicle is assigned a route to follow. Similarly, existing works use waypoints \cite{gulino2023waymax, suo2023mixsim} or goal points \cite{kazemkhani2024gpudrive, apple2025self-play}. Vehicles terminate upon colliding or leaving the road.

\begin{figure}[t]
    \centering
    \vspace{4pt}
    \includegraphics[width=0.42\textwidth]{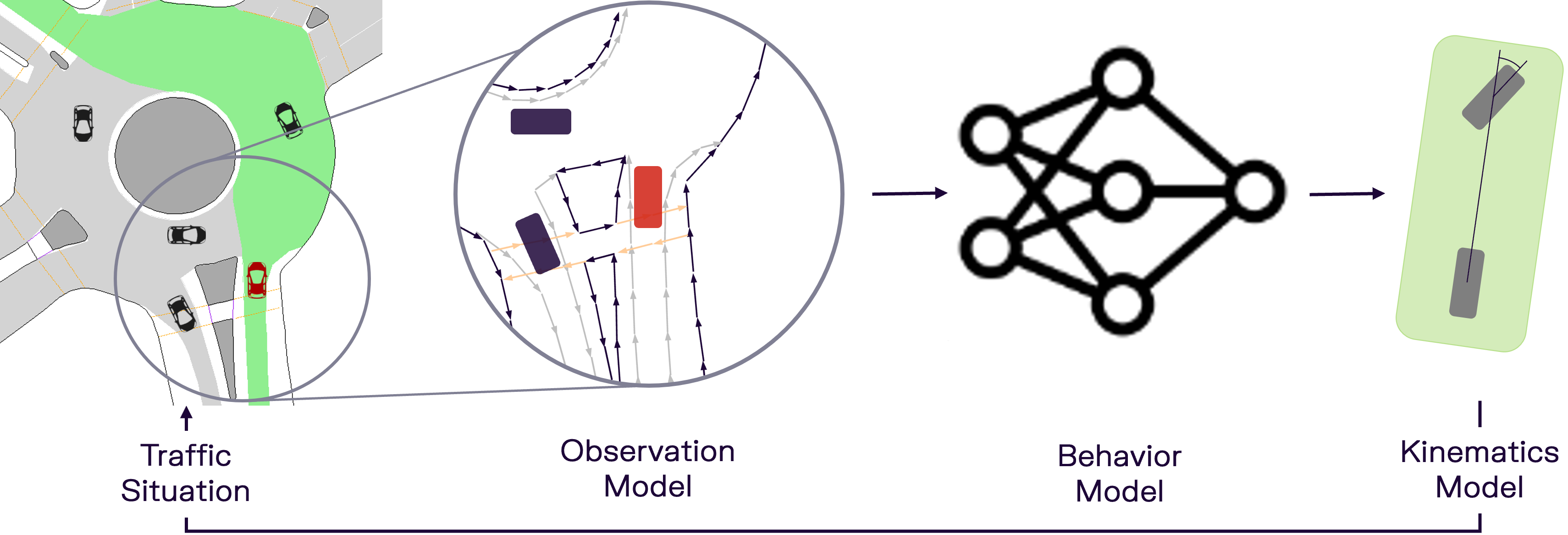}
    \vspace{-10pt}
    \caption{Single simulation step: The behavior model maps observations to actions, which are executed via a kinematic bicycle model for vehicles and kinematic unicycle model for pedestrians.}
    \label{fig:sim_setup}
    \vspace{-5pt}
\end{figure}

As illustrated in Figure \ref{fig:sim_setup}, simulations evolve recursively. First, an observation model encodes the traffic situation, either in an agent-centric or instance-centric manner. Next, a learned behavior model maps the observation to actions, \ie acceleration and steering angle for vehicles, and acceleration and heading rate for pedestrians. Finally, these actions are executed via a kinematics model, updating the agent's state. Repeating these steps for all agents at fine-grained time intervals yields continuous traffic simulation. Our in-house simulation backbone is implemented in C++, thus achieving a low runtime. Figure \ref{fig:example_sit} shows an example scenario.

This sequential decision-making process is formulated as a \gls{pomdp}, characterized by the tuple $(S, O, A, T, R, \Omega, \gamma)$. In this framework, the agent does not have direct access to the true state $s \in S$ of the environment. Instead, it receives a noisy observation $o \in O$, provided by the observation model $\Omega: S \to O$. Upon executing an action $a \in A$, the state of the environment is updated stochastically according to the transition probability density $T: S \times A \times S \rightarrow \left[ 0, \infty \right)$. In addition, the agent receives a numerical reward defined by the reward function $R: S \times A \to \mathbb{R}$, as well as a new observation of the updated environment state. The discount factor $\gamma \in \left[0, 1\right)$ balances the trade-off between immediate and future rewards. The solution to a \gls{pomdp} is the optimal policy $\pi^*: O \times A \to \left[0, \infty \right)$, which maps observations to action distributions, maximizing the expected cumulative reward $J(\pi) = \mathop{\mathbb{E}}_{\pi} \left[ \sum_{k=0}^\infty \gamma^k r_k \right]$, where $r_k = R(s_k, a_k)$.

\subsection{Reinforcement Learning}
We maximize $J(\pi)$ using \gls{rl}, where an agent learns effective actions by interacting with a simulated environment. The process alternates between 1) collecting a set of experiences $E = \{ e_1, \dots, e_M \}$, with each $e_k = (o_k, a_k, r_k)$ representing a single-step experience sample, and 2) updating the policy to reinforce actions leading to higher rewards. These steps are alternated until a sufficiently good policy is found.

Commonly, the parameterized policy $\pi_\theta$ is updated via:
\begin{equation}
    \theta \leftarrow \theta + \alpha \frac{1}{M} \sum_{e_k \in E} A(o_k, a_k) \triangledown_\theta \log \pi_\theta (a_k \mid o_k),
\end{equation}
where $\alpha$ is the learning rate and the gradients of the policy $\triangledown_\theta \log \pi_\theta(a_k \mid o_k)$ are weighted by the advantage $A(o_k, a_k)$.

In our multi-agent setting, we apply a shared policy across all agents, trained via self-play \gls{rl}. Specifically, we use \gls{gae} \cite{Schulmann15} for estimating $A(o_k, a_k)$ and \gls{ppo} \cite{Schulmann17} for optimizing the policy. We refer to \cite{apple2025self-play} for a detailed description of how self-play \gls{rl} is applied to the task of learning behavior models for simulating driver behavior.

\subsection{Adversarial Inverse Reinforcement Learning} \label{sec:airl}
\gls{rl} requires defining a reward function that accurately captures the incentive structure of real-world driving. Instead of manually defining such a reward function, we use \gls{airl} \cite{fu2018airl} to reconstruct a surrogate reward signal from real data, given as $\mathcal{D}=\{ (o_1, a_1), (o_2, a_2), \dots \}$. In \gls{airl}, an additional discriminator model $D_\phi$ is trained to distinguish generated from real samples, producing the probability $D_\phi(o, a) \in \left[0, 1 \right]$ for the observation-action pair being real, \ie stemming from $\mathcal{D}$. The policy is trained via \gls{rl} using the surrogate reward
\begin{equation}\label{eq:airl_surrogate}
    \tilde{r}(o, a) = \log D_\phi(o, a) - \log \big( 1 - D_\phi(o, a) \big),
\end{equation}
which guides the policy toward taking realistic actions. Following \cite{Konstantinidis2024IV}, to smooth the decision boundary during discriminator training, we add random noise to the actions in $\mathcal{D}$, matching the standard deviation of the policy.

\newcommand{\rotimg}[2]{%
\begin{tikzpicture}[baseline]
\node[inner sep=0pt, outer sep=0pt] (img)
{\rotatebox{90}{\includegraphics[width=0.1464\textwidth]{#1}}};
\node[anchor=north west, inner sep=2pt, outer sep=0pt] at (img.north west) {\tiny #2};
\end{tikzpicture}%
}
\begin{figure}[t]
\centering
\rotimg{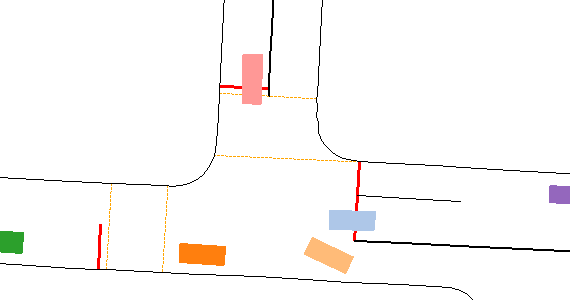}{$t\!=\!\SI{0}{\second}$}\hfill
\rotimg{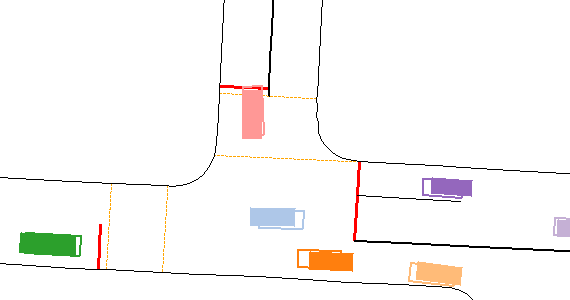}{$t\!=\!\SI{2}{\second}$}\hfill
\rotimg{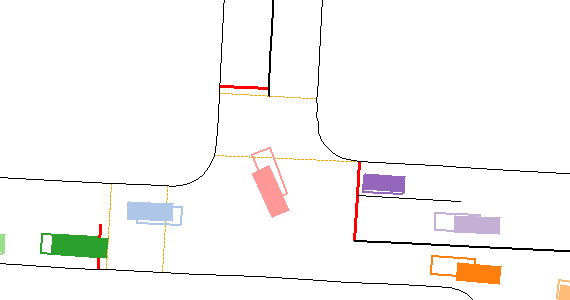}{$t\!=\!\SI{4}{\second}$}\hfill
\rotimg{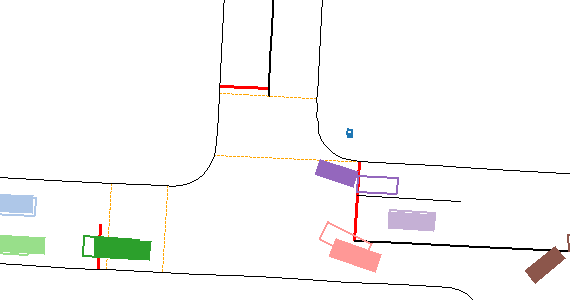}{$t\!=\!\SI{6}{\second}$}\hfill
\rotimg{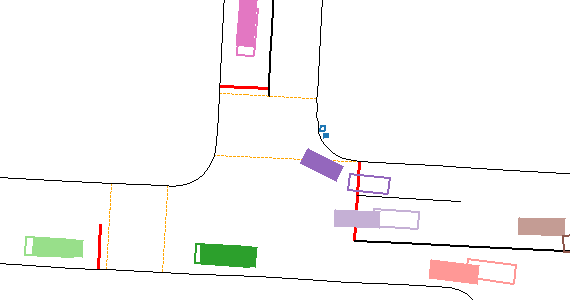}{$t\!=\!\SI{8}{\second}$}\hfill
\rotimg{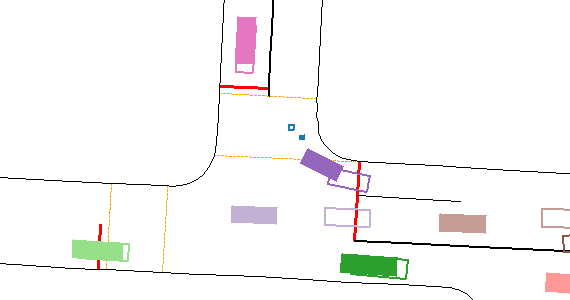}{$t\!=\!\SI{10}{\second}$}
\vspace{-2pt}
\caption{Illustration of an example scenario using instance-centric observations. Simulated vehicles and their counterparts from the real-world dataset are depicted as solid and outlined rectangles, respectively.}
\label{fig:example_sit}
\vspace{-15pt}
\end{figure}

\begin{figure*}[t]
  \centering
  \vspace{4pt}
  \includegraphics[width=0.9\textwidth]{images/model_ICRA_v2.png}
  \vspace{-8pt}
  \caption{Illustration of the proposed instance-centric behavior model mapping observations to actions. Instance encoders convert observations into latent tokens. These tokens are then augmented with positional encodings relative to the target agent before being passed through multiple layers of our refinement module. Lastly, each refined agent token is decoded into its corresponding actions. Static map tokens can be reused across simulation steps.}
  \label{fig:model}
  \vspace{-14pt}
\end{figure*}

\textbf{Adaptive Reward Transformation}: 
Typically, the discriminator classifies samples correctly, leading to a negative expected value for $\tilde{r}(o, a)$. This often drives agents to terminate as fast as possible to avoid constantly accumulating penalties. To mitigate this, in \cite{Konstantinidis2024IV} a positive constant $c=5$ is added to $\tilde{r}$, shifting its expected value into the positive range. Intuitively, this adjustment encourages survival without altering the policy's optimal behavior \wrt the discriminator.

However, using a fixed $c$ across experiments, \eg when comparing different discriminator architectures or scene representations, may result in varying expected values of $\tilde{r}$, thereby influencing the agent's incentive to survive. For instance, a weaker discriminator tends to learn a less distinct decision boundary, yielding higher surrogate rewards, whereas a stronger discriminator learns sharper boundaries, producing lower surrogate rewards. Consequently, policies trained with stronger discriminators have less incentive to avoid termination, \ie collisions or off-track driving, despite in both cases the optimal behavior \wrt the discriminator is to match the distribution of the real data $\mathcal{D}$.

To decouple the agent's survival incentive from the discriminator's accuracy, we propose an adaptive reward offset
\begin{equation}\label{eq:target_mean}
    c(n)=\tilde{r}_\mathrm{target} - \tilde{r}_\mathrm{mean}(n),
\end{equation}
where $\tilde{r}_\mathrm{mean}(n)$ is the average surrogate reward for the generated samples in epoch $n$, and $\tilde{r}_\mathrm{target}$ is a hyperparameter, specifying the desired average reward. In other words, our dynamic reward offset can be interpreted as an additional reward term that encourages the policy to survive longer, especially when the discriminator is confident that the given samples are generated. This adaptive approach keeps the expected value of $\tilde{r}$ consistent, thus enabling fair comparisons across experiments, while still allowing the discriminator to guide the policy toward more realistic behavior by computing the appropriate offset to $\tilde{r}_\mathrm{target}$.

Technically, our adaptive reward transformation is a form of reward shaping. However, unlike conventional shaping methods that add a fixed term to guide the policy toward desirable behavior, our offset adapts dynamically to the discriminator’s performance, automatically balancing realism and robustness during training.

\subsection{Instance-Centric Observations}
Rather than adopting agent-centric observations, as prior works do, \eg \cite{bhattacharyya2022modeling, apple2025self-play, cornelisse2025self-play}, we employ instance-centric observations, representing each instance, \ie the $N_a$ agents and $N_p$ map elements, in its respective local coordinate frame. 

For static map elements, such as pedestrian crossings or lane boundaries, we use a vectorized representation \cite{vectornet}. Each map element is approximated by a polyline with a maximum total length of \SI{10}{m}. A polyline consists of multiple vectors, each described by a set of features $\mathbf{v} = \left[ \mathbf{v}_\mathrm{start}, \mathbf{v}_\mathrm{end}, \mathbf{v}_\mathrm{type} \right]^\intercal$, which includes the vector's start and end points, as well as a one-hot encoding for the polyline type. For each polyline, the origin of its instance-centric coordinate frame is defined as the mean position of all polyline points. The x-axis is aligned with the mean direction of the vectors forming the polyline.

The local coordinate frame of an agent is determined by its last position and moving direction. Agents are described by their size (width and length), velocity, current speed limit, and a binary indicator for \glspl{vru}: $\mathbf{x} = \left[ \mathbf{x}_\mathrm{size}, x_\mathrm{vel}, x_\mathrm{limit}, x_\mathrm{vru} \right]^\intercal$. Note that, compared to existing works utilizing agent-centric observations, here the agent's position and orientation are not included in the agent features, but are now captured by the position and orientation of its local coordinate frame. 

Intuitively, the local coordinate frames serve as anchor poses for the individual instances, with $\mathbf{p}_i$ being the origin and $\alpha_i$ being the heading of the $i$-th instance relative to an arbitrary global coordinate frame. Following \cite{zhang2024simpl, shi2024mtr++}, to capture the spatial relationship between two instances $i$ and $j$, we define a coordinate transformation from the local frame of $i$ to that of $j$. Specifically, the relative pose of instance $j$ in the frame of instance $i$ is represented as 
\begin{equation}
    \mathbf{r}_{i \rightarrow j} = \left[ \Delta \alpha_{i \rightarrow j}, \psi_{i \rightarrow j}, \left\| \mathbf{p}_{i \rightarrow j} \right\| \right]^\intercal,
\end{equation}
where $\Delta \alpha_{i \rightarrow j} = \alpha_j - \alpha_i$ denotes the heading difference, $\psi_{i \rightarrow j}$ is the relative azimuth, \ie the angle of $\mathbf{p}_{i \rightarrow j} = \mathbf{p}_j - \mathbf{p}_i$ in the reference frame of $i$, and $\left\| \mathbf{p}_{j \rightarrow i} \right\|$ is the distance between the two anchor poses. Both angles are provided as unit-circle embeddings $f(x)=(\cos x, \sin x)$.

\subsection{Behavior Model Details}\label{sec:model}
After obtaining a new observation, it needs to be mapped to the actions that should be executed upon making that observation. Figure \ref{fig:model} illustrates our proposed model.

\textbf{\emph{Instance Feature Encoders}}:
We begin by processing the observation into a set of latent instance-tokens. Following the approach introduced in \cite{vectornet}, the $L$ vectors of a polyline are encoded through multiple layers of message-passing:
\begin{equation}\label{eq:static_context_mp}
    \mathbf{v} \leftarrow f_\mathrm{rel} \left(g_\mathrm{enc} \left( \mathbf{v} \right), f_\mathrm{agg} \left( \{g_\mathrm{enc} \left( \mathbf{v}_l \right)\}_{l=1}^L \right) \right),
\end{equation} 
followed by an element-wise max-pooling operation over $\{\mathbf{v}_1, \mathbf{v}_2, \dots, \mathbf{v}_L\}$ to produce the final polyline token $\mathbf{z}^{\mathrm{(map)}}$. We implement $g_\mathrm{enc}(\cdot)$ as a \gls{mlp}, $f_\mathrm{agg}(\cdot)$ as element-wise max-pooling, and $f_\mathrm{rel}(\cdot)$ as concatenation. Conversely, each agent feature vector $\mathbf{x}$ is encoded into the same latent space using another \gls{mlp}, producing the agent token $\mathbf{z}^{\mathrm{(agent)}} = \mathrm{MLP}(\mathbf{x})$. 

Note that using agent-centric observations, requires each instance to be encoded $N_a \times H$ times — once per agent per simulation step, where $H$ denotes the total number of steps. In contrast, with our instance-centric representation, the polyline tokens $Z^{\mathrm{(map)}} = \{\mathbf{z}^{\mathrm{(map)}}_i \mid i = 1, \cdots, N_p\}$ remain unchanged throughout the rollout. This enables reusing the tokens in subsequent simulation steps, reducing the number of encoding a polyline-instance from $N_a \times H$ times to just one. However, agent states evolve during the simulation rollouts, requiring agent tokens $Z^{\mathrm{(agent)}} = \{\mathbf{z}^{\mathrm{(agent)}}_i \mid i = 1, \cdots, N_a\}$ to be regenerated at each simulation step.
Nonetheless, since agents are encoded in a viewpoint-invariant manner, they only must be encoded $H$ times instead of $N_a \times H$ times.

\textbf{\emph{Query-Centric Symmetric Context Encoder}}:
After obtaining the instance tokens, we use a symmetric context encoder to refine the queried agent instance tokens in a viewpoint-invariant manner. The idea is to model the interactions between the target agent and nearby instance tokens. 

For that, we define the relative positional encoding between a target agent $i \in \{1, \cdots, N_a\}$ and any of the combined instance tokens $Z = \left[ Z^{\mathrm{(agent)}},\; Z^{\mathrm{(map)}} \right]$, indexed by $j$, as $\mathbf{c}_{i \rightarrow j} = \mathbf{r}_{i \rightarrow j} \oplus \mathds{1}^{(\mathrm{agent})}_j \oplus \mathds{1}^{(\mathrm{route})}_j$, where $\oplus$ denotes concatenation. The binary indicators $\mathds{1}^{(\mathrm{agent})}_j$ and $\mathds{1}^{(\mathrm{route})}_j$ specify whether instance $j$ is an agent, or a map polyline that is part of the route that agent $i$ should follow, respectively. These indicators are appended here rather than to the vector features $\mathbf{v}$, as each agent may follow a different route, making a global encoding at the vector level infeasible. 
Finally, following the approach proposed in \cite{perez2018film}, we compute pairwise encodings by applying a feature-wise linear transformation:
\begin{equation}
    \mathbf{z}_{i \rightarrow j} = \zeta\left(\mathbf{c}_{i \rightarrow j} \right) \odot \mathbf{z}_j + \beta\left(\mathbf{c}_{i \rightarrow j}\right),
\end{equation}
where $\zeta(\cdot)$ and $\beta(\cdot)$ are both realized as \gls{mlp}s. This operation yields the representation of token $\mathbf{z}_j$ relative to instance $i$.

Similarly to \cite{chen2024drivingwithllms, zhang2024simpl}, we then refine the instance tokens of simulated target agents through multiple Perceiver layers \cite{jaegle2021perceiver}. Each layer consists of a \gls{mhca} module with a skip connection, followed by layer normalization, an \gls{mlp}, also with a skip connection, and another layer normalization. The \gls{mhca} is expressed as:
\begin{equation}\label{eq:mhca}
    \mathbf{z}^{'(k)}_{i} = \mathrm{MHCA}^{(k)} \left(\mathrm{Q}\!: \mathbf{z}^{(k-1)}_{i}, \textrm{ } \mathrm{KV}\!: \{\mathbf{z}_{i \rightarrow j}\}_{j \in \Gamma(i)} \right),
\end{equation}
where $k \in \{1, \cdots, K\}$ denotes the $k$-th layer, and $\Gamma(i)$ is the set of neighboring token indices for agent $i$, limited to those with anchor points within a set observation radius around agent $i$ to ensure efficiency. We initialize the first layer with $\mathbf{z}_{i}^{(0)} = \mathbf{z}_{i \rightarrow i}$ and use the output of layer $k$, denoted $\mathbf{z}_{i}^{(k)}$, as input to the subsequent Perceiver layer.

\textbf{\emph{Decoder}}:
The decoder processes the refined instance token $\mathbf{z}_{i}^{(K)}$ from the last Perceiver layer. In the behavior model, it outputs the mean and standard deviation of the next action. In the discriminator, which evaluates observation-action pairs, the action is appended to $\mathbf{z}_{i}^{(K)}$ to then compute the classification score $D_\phi$.

We use three layers of the message passing mechanism defined in \eqref{eq:static_context_mp} and $16$ channels per head in \eqref{eq:mhca}. Each \gls{mlp} consists of a linear layer, followed by layer normalization, ReLU activation, and a second linear layer.

\section{Experiments}

\textbf{\emph{Datasets}}:
Evaluation is conducted on two datasets: 1) the publicly available INTERACTION dataset \cite{zhan2019interaction}, containing \num{38255} trajectories (\SI{4.15}{\%} \glspl{vru}) across \num{11} locations; and 2) an in-house dataset commercially licensed by DeepScenario, comprising \num{88540} trajectories (\SI{43.59}{\%} \glspl{vru}) across \num{8} locations. Both datasets were recorded using drones, ensuring high-quality data, and focus on interactive scenarios such as merging lanes, unsignalized intersections, and roundabouts. We use \SI{20}{\%} of the recordings per location for validation and \SI{30}{\%} for testing. A more compact version of the DeepScenario dataset \cite{dhaouadi2025deepscenario} is publicly available and yields similar results.

\textbf{\emph{Baselines}}:
We compare against several compact baselines:
\begin{enumerate}
\item \emph{\gls{cv}}: A learning-free baseline that assumes agents continue forward at a constant velocity.
\item \emph{LateFusionMLP} \cite{cornelisse2025self-play}: Following \cite{apple2025self-play, cornelisse2025self-play, kazemkhani2024gpudrive}, this compact agent-centric model consists solely of \gls{mlp}s and max-pooling operations. We adopt the public implementation \cite{cornelisse2025self-play}, replacing its discrete action decoder with ours to support continuous actions and training it within our framework for realistic behavior modeling.
\item \emph{GraphAIRL} \cite{Konstantinidis2024IV}: A more sophisticated agent-centric model that uses a vectorized scene representation \cite{vectornet} and attention-based interaction modeling. We evaluate two variants: 1) trained with $c=5$, as done in \cite{Konstantinidis2024IV}, and 2) trained with our proposed adaptive reward offset \eqref{eq:target_mean}.
\item \emph{Behavior Cloning (\gls{bc})}: A supervised learning variant of our instance-centric approach, trained for \num{600} epochs by minimizing the negative log-likelihood of expert actions under the predicted action distribution.
\end{enumerate}
Our agent-centric observations include both nearby agents and map elements within the observation radius. The start and end points of a vector $\mathbf{v}$ are expressed in the target agent’s coordinate frame, with the route indicator $\mathds{1}^{(\mathrm{route})}_j$ concatenated to $\mathbf{v}$. Positions and orientations of surrounding agents, relative to the target agent, are appended to $\mathbf{x}$.

\textbf{\emph{Training Details}}:
All models are trained using the AdamW optimizer \cite{loshchilov2017adamw} with a batch size of $1024$. Learning rates are set to $0.0002$ for the policy model and $0.0001$ for the discriminator, both decayed by a factor of $10$ during the final \SI{30}{\%} of training. For \gls{rl}, we use advantage normalization. The models are trained for \num{10000} epochs with a discount factor $\gamma = 0.95$, \gls{gae} parameter $\lambda = 0.95$, and target rewards $\tilde{r}_\mathrm{target}=33$ and $\tilde{r}_\mathrm{target}=9$ for INTERACTION and DeepScenario, respectively. For the policy, the observation radius is set to $\SI{50}{m}$. Since the number of observed instances grows quadratically with the observation range, we reduce the discriminator's observation radius to $\SI{30}{m}$, allowing it to focus primarily on the target agent and its direct surroundings.

In each epoch, scenarios with approximately \num{880} agents are sampled from the training set. Validation scenarios contain roughly \num{1650} agents, while test scenarios feature \num{5109} and \num{19632} agents for INTERACTION and DeepScenario, respectively. Agents are uniformly distributed across locations. Each scenario is simulated for \SI{10}{s} at \SI{5}{Hz}. 
For final evaluation, we select the model that achieves the lowest aggregated score
\begin{equation}
    S(\theta) = \frac{
        \sqrt{\frac{1}{N_a} \sum_{n=1}^{N_a} \left(\mathrm{FDE}_n^{(\pi_\theta)}\right)^2}
    }{
            \max (1 - \mathrm{off}^{(\pi_\theta)} - \mathrm{col}^{(\pi_\theta)},\; \epsilon )
    },
\end{equation}
\ie the root mean squared final displacement error normalized by $(1 - \text{Off-Track Rate} - \text{Collision Rate})$. To ensure numerical stability, we set $\epsilon = 10^{-6}$. The \gls{fde} is the Euclidean distance to the ground truth agent at the final simulation step. This score balances positional accuracy and robustness. 
During evaluation, terminated vehicles are not removed from simulation.

We use three Perceiver layers and a hidden dimension of $128$. Additionally, we evaluate a smaller variant with one Perceiver layer and a hidden dimension of $64$, denoted \textit{Ours (small)}. All baseline models use a hidden dimension of $128$.

\begin{table*}[!h]
    \centering
    \vspace{4pt}
    \renewcommand{\arraystretch}{1.2} 
    \caption{Model performance on the deepscenario dataset.}
    \vspace{-8pt}
    \resizebox{\textwidth}{!}{%
    \begin{tabular}{c c c | c c c c c c c c | c c }
    \hline
    \multirow{2}{*}{\textbf{Model}} &
    \textbf{Observation} &
    \textbf{Num} &
    \iffde \multicolumn{3}{c|}{\textbf{FDE}} \else \multicolumn{3}{c|}{\textbf{RMSE}} \fi &
    \multicolumn{2}{c|}{\textbf{Collision}$^{\dagger}$ \textbf{w/}} &
    \multicolumn{2}{c|}{\textbf{Off-}$^{\dagger}$} &
    \textbf{Aggregated} &
    \multicolumn{2}{c}{\textbf{Training Time}} \\
    & \textbf{Type}
    & \textbf{Parameter}
    & \textbf{All} & \textbf{non-VRU} & \textbf{VRU}
    & \textbf{non-VRU} & \textbf{VRU}
    & \textbf{Route} & \textbf{Track}
    & \textbf{Score} 
    & \textbf{T4 16GB} 
    & \textbf{A100 80GB} \\
    \hline
    Expert Demo & 
        - &
        - &
        $0.0$ & 
        $0.0$ & 
        $0.0$ & 
        $0.16$ & 
        $0.05$ & 
        $3.75$ & 
        $0.83$ & 
        $0.0$ &
        - &
        - \\
    \hline
    CV & 
        - &
        - &
        \iffde
            $16.73$ & 
            $29.18$ & 
            $\mathbf{3.99}$ & 
        \else
            $27.78$ & 
            $38.44$ & 
            $\underline{6.97}$ & 
        \fi
        $26.16$ & 
        $3.76$ & 
        $57.52$ & 
        $49.17$ & 
        $123.72$ &
        - &
        - \\
    LateFusionMLP$^{\ddagger}$ \cite{cornelisse2025self-play} & 
        AC &
        $105$k &
        \iffde
            $9.28_{\pm0.14}$ & 
            $13.10_{\pm0.23}$ & 
            $5.33_{\pm0.20}$ & 
        \else
            $14.19_{\pm0.23}$ & 
            $17.84_{\pm0.39}$ & 
            $8.97_{\pm0.39}$ & 
        \fi
        $1.61_{\pm0.23}$ & 
        $0.55_{\pm0.09}$ & 
        $3.89_{\pm0.56}$ & 
        $\underline{0.61}_{\pm0.02}$ & 
        $14.59_{\pm0.22}$ &
        $86.85$ &
        $40.75$ \\
    GraphAIRL \cite{Konstantinidis2024IV} & 
        AC &
        $146$k &
        \iffde
            $7.86_{\pm0.15}$ & 
            $10.78_{\pm0.18}$ & 
            $4.83_{\pm0.24}$ & 
        \else
            $12.26_{\pm0.28}$ & 
            $15.25_{\pm0.38}$ & 
            $8.04_{\pm0.40}$ & 
        \fi
        $1.09_{\pm0.38}$ & 
        $0.79_{\pm0.26}$ & 
        $3.96_{\pm0.33}$ & 
        $0.70_{\pm0.13}$ & 
        $12.58_{\pm0.28}$ &
        $125.94$ &
        $45.95$ \\
    GraphAIRL$^{\ddagger}$ \cite{Konstantinidis2024IV} & 
        AC &
        $146$k &
        \iffde
            $7.79_{\pm0.12}$ & 
            $10.84_{\pm0.19}$ & 
            $4.60_{\pm0.14}$ & 
        \else
            $12.20_{\pm0.20}$ & 
            $15.28_{\pm0.29}$ & 
            $7.76_{\pm0.31}$ & 
        \fi
        $0.67_{\pm0.27}$ & 
        $0.42_{\pm0.29}$ & 
        $4.16_{\pm0.55}$ & 
        $\underline{0.61}_{\pm0.03}$ & 
        $12.41_{\pm0.24}$ &
        $129.03$ &
        $47.10$ \\
    BC & 
        IC &
        $430$k &
        \iffde
            $7.75_{\pm0.53}$ & 
            $11.40_{\pm0.97}$ & 
            $\underline{4.03}_{\pm0.25}$ & 
        \else
            $12.07_{\pm0.70}$ & 
            $15.58_{\pm1.04}$ & 
            $\mathbf{6.82}_{\pm0.29}$ & 
        \fi
        $14.71_{\pm2.92}$ & 
        $2.90_{\pm0.28}$ & 
        $16.72_{\pm10.43}$ & 
        $10.51_{\pm9.80}$ & 
        $17.35_{\pm5.22}$ &
        $12.67$ &
        $3.07$ \\
    \hline
    Ours (small) & 
        IC &
        $60$k &
        \iffde
            $\underline{7.59}_{\pm0.23}$ & 
            $\underline{10.31}_{\pm0.35}$ & 
            $4.74_{\pm0.19}$ & 
        \else
            $\underline{11.83}_{\pm0.30}$ & 
            $\underline{14.65}_{\pm0.43}$ & 
            $7.85_{\pm0.23}$ & 
        \fi
        $\underline{0.66}_{\pm0.33}$ & 
        $\underline{0.27}_{\pm0.09}$ & 
        $\underline{3.85}_{\pm0.70}$ & 
        $\mathbf{0.57}_{\pm0.03}$ & 
        $\underline{12.01}_{\pm0.34}$ &
        $45.53$ &
        $25.13$ \\
    Ours & 
        IC &
        $430$k &
        \iffde
            $\mathbf{7.09}_{\pm0.23}$ & 
            $\mathbf{9.69}_{\pm0.24}$ & 
            $4.43_{\pm0.26}$ & 
        \else
            $\mathbf{11.17}_{\pm0.33}$ & 
            $\mathbf{13.86}_{\pm0.40}$ & 
            $7.45_{\pm0.32}$ & 
        \fi
        $\mathbf{0.38}_{\pm0.13}$ & 
        $\mathbf{0.14}_{\pm0.05}$ & 
        $\mathbf{3.27}_{\pm0.23}$ & 
        $\mathbf{0.57}_{\pm0.03}$ & 
        $\mathbf{11.29}_{\pm0.35}$ &
        $125.75$ &
        $44.07$ \\
    \hline
    \end{tabular}
    }
    \vspace{-3pt}
    {\parbox{0.99\linewidth}{
        \vspace{1pt}
        \scriptsize
        Reported as mean $\pm$ std across $7$ random seeds. '${\dagger}$' indicates metric reported for the non-VRU subset. '${\ddagger}$' indicates baseline model trained with the proposed target reward. FDE in [m]. Collision, off-route and off-track rates in [\%]. Training times in [h]. AC = agent-centric. IC = instance-centric.
    }}
    \label{tab:model_comparison_DEEPSCENARIO}
    \vspace{-8pt}
\end{table*}

\begin{table}[!h]
    \centering
    \renewcommand{\arraystretch}{1.2} 
    \caption{Model performance on the interaction dataset.}
    \vspace{-8pt}
    \resizebox{0.98\columnwidth}{!}{%
    \begin{tabular}{c | c c c c | c c}
    \hline
    \multirow{2}{*}{\textbf{Model}} &
    \iffde \multirow{2}{*}{\textbf{FDE}} \else \multirow{2}{*}{\textbf{RMSE}} \fi &
    \multirow{2}{*}{\textbf{Collision}$^{\dagger}$} &
    \multirow{2}{*}{\textbf{Off-Track}$^{\dagger}$} &
    \textbf{Aggregated} & 
    \multicolumn{2}{c}{\textbf{Training Time}} \\
    & & & & \textbf{Score} & \textbf{T4 16GB} & \textbf{A100 80GB} \\
    \hline
    Expert Demo & 
        $0.0$ & 
        $0.0$ & 
        $0.51$ &
        $0$ &
        - &
        - \\
    \hline
    CV & 
        \iffde
            $17.44$ & 
        \else
            $23.17$ & 
        \fi
        $22.59$ & 
        $30.65$ &
        $49.55$ &
        - &
        - \\
    LateFusionMLP$^{\ddagger}$ \cite{cornelisse2025self-play} & 
        \iffde
            $10.12_{\pm0.14}$ & 
        \else
            $14.08_{\pm0.30}$ & 
        \fi
        $0.57_{\pm0.24}$ & 
        $\mathbf{0.30}_{\pm0.02}$ &
        $14.20_{\pm0.30}$ &
        $97.57$ &
        $44.48$ \\
    GraphAIRL \cite{Konstantinidis2024IV} & 
        \iffde
            $\underline{8.51}_{\pm0.35}$ & 
        \else
            $\underline{12.27}_{\pm0.43}$ & 
        \fi
        $1.38_{\pm0.48}$ & 
        $0.36_{\pm0.04}$ &
        $\underline{12.49}_{\pm0.48}$ &
        $152.35$ &
        $52.26$ \\
    GraphAIRL$^{\ddagger}$ \cite{Konstantinidis2024IV} & 
        \iffde
            $9.05_{\pm0.47}$ & 
        \else
            $12.81_{\pm0.48}$ & 
        \fi
        $0.70_{\pm0.25}$ & 
        $\mathbf{0.30}_{\pm0.03}$ &
        $12.94_{\pm0.50}$ &
        $157.19$ &
        $55.14$ \\
    BC & 
        \iffde
            $10.28_{\pm0.51}$ & 
        \else
            $14.45_{\pm0.51}$ & 
        \fi
        $13.98_{\pm2.15}$ & 
        $9.65_{\pm7.02}$  &
        $19.09_{\pm2.22}$ &
        $2.37$ &
        $0.68$ \\
    \hline
    Ours (small) & 
        \iffde
            $8.78_{\pm0.39}$ & 
        \else
            $12.47_{\pm0.59}$ & 
        \fi
        $\underline{0.40}_{\pm0.22}$ & 
        $\underline{0.32}_{\pm0.04}$ &
        $12.56_{\pm0.58}$ &
        $42.47$ &
        $24.17$ \\
    Ours & 
        \iffde
            $\mathbf{8.34}_{\pm0.29}$ & 
        \else
            $\mathbf{11.99}_{\pm0.27}$ & 
        \fi
        $\mathbf{0.38}_{\pm0.18}$ & 
        $\mathbf{0.30}_{\pm0.02}$ &
        $\mathbf{12.08}_{\pm0.26}$ &
        $126.08$ &
        $50.43$ \\
    \hline
    \end{tabular}
    }
    \vspace{-3pt}
    {\parbox{0.96\linewidth}{
        \vspace{1pt}
        \scriptsize
         Reported as mean $\pm$ std across $7$ random seeds.
    }}
    \label{tab:model_comparison_INTERACTION_IN_Dataset}
\end{table}

\begin{table}[!h]
    \centering
    \renewcommand{\arraystretch}{1.2} 
    \caption{Cross-Dataset performance on the interaction dataset.}
    \vspace{-10pt}
    \resizebox{0.78\columnwidth}{!}{%
    \begin{tabular}{c | c c c c }
    \hline
    \multirow{2}{*}{\textbf{Model}} &
    \iffde \multirow{2}{*}{\textbf{FDE}} \else \multirow{2}{*}{\textbf{RMSE}} \fi &
    \multirow{2}{*}{\textbf{Collision}$^{\dagger}$} &
    \multirow{2}{*}{\textbf{Off-Track}$^{\dagger}$} &
    \textbf{Aggregated} \\
    & & & & \textbf{Score} \\
    \hline
    Expert Demo & 
        $0.0$ & 
        $0.0$ & 
        $0.51$ &
        $0.0$ \\
    \hline
    CV & 
        \iffde
            $17.44$ & 
        \else
            $23.17$ & 
        \fi
        $22.59$ & 
        $30.65$ &
        $49.55$ \\
    LateFusionMLP$^{\ddagger}$ \cite{cornelisse2025self-play} & 
        \iffde
            $15.22_{\pm1.34}$ & 
        \else
            $19.65_{\pm1.77}$ & 
        \fi
        $6.47_{\pm1.97}$ & 
        $6.89_{\pm2.85}$ &
        $22.82_{\pm3.21}$ \\
    GraphAIRL \cite{Konstantinidis2024IV} & 
        \iffde
            $16.05_{\pm1.14}$ & 
        \else
            $20.51_{\pm1.54}$ & 
        \fi
        $5.54_{\pm2.57}$ & 
        $4.12_{\pm2.13}$ &
        $22.80_{\pm2.62}$ \\
    GraphAIRL$^{\ddagger}$ \cite{Konstantinidis2024IV} & 
        \iffde
            $14.44_{\pm0.49}$ & 
        \else
            $18.94_{\pm0.71}$ & 
        \fi
        $3.24_{\pm0.84}$ & 
        $3.89_{\pm3.06}$ &
        $20.43_{\pm1.44}$ \\
    BC & 
        \iffde
            $15.62_{\pm1.31}$ & 
        \else
            $20.04_{\pm1.32}$ & 
        \fi
        $19.69_{\pm1.46}$ & 
        $13.95_{\pm3.65}$ &
        $30.25_{\pm2.15}$ \\
    \hline
    Ours (small) & 
        \iffde
            $\underline{14.26}_{\pm1.48}$ & 
        \else
            $\underline{18.57}_{\pm1.67}$ & 
        \fi
        $\underline{0.81}_{\pm0.28}$ & 
        $\underline{1.49}_{\pm0.79}$ &
        $\underline{19.01}_{\pm1.72}$ \\
    Ours & 
        \iffde
            $\mathbf{12.62}_{\pm0.69}$ & 
        \else
            $\mathbf{16.55}_{\pm0.68}$ & 
        \fi
        $\mathbf{0.51}_{\pm0.16}$ & 
        $\mathbf{1.11}_{\pm0.16}$ &
        $\mathbf{16.82}_{\pm0.71}$ \\
    \hline
    \end{tabular}
    }
    \vspace{-3pt}
    {\parbox{0.78\linewidth}{
        \vspace{1pt}
        \scriptsize
         Reported as mean $\pm$ std across $7$ random seeds. Models are trained on DeepScenario dataset and evaluated on INTERACTION dataset. 
    }}
    \label{tab:model_comparison_INTERACTION_CROSS_Dataset}
    \vspace{-13pt} 
\end{table}

\subsection{Results}

We evaluate our proposed approach against a diverse set of baselines on two automated driving datasets. Table \ref{tab:model_comparison_DEEPSCENARIO} presents the quantitative results on the DeepScenario dataset. 
The proposed model achieves the lowest \gls{fde} for the overall and non-\gls{vru} subset, as well as the most robust behavior in terms of collision, off-route and off-track rates. Overall, our model outperforms all baselines on seven out of eight metrics, and even our small variant with only \num{60}k parameters surpasses all baselines on the same seven metrics. The only exception is \gls{vru} \gls{fde}, where two baselines achieve marginally better positional accuracy.
Notably, our approach achieves lower off-route and off-track rates than ground truth agents, whose errors largely stem from annotation noise.

The learning-free \emph{\gls{cv}} baseline achieves reasonable positional accuracy for \glspl{vru}, as most do not abruptly change speed or direction, but performs poorly on other metrics, underscoring the need to model interactions. \emph{LateFusionMLP} captures more realistic behaviors but still suffers from low robustness and comparably high \gls{fde}, likely due to limited model capacity. Both \emph{GraphAIRL} variants demonstrate high positional accuracy and robustness. Notably, training the model with our adaptive reward offset significantly improves its robustness and yields a small gain in positional accuracy. Finally, the \emph{\gls{bc}} baseline achieves low \gls{fde} but incurs extremely high off-track and collision rates due to the aforementioned covariate shift when executed in simulation.

Table \ref{tab:model_comparison_INTERACTION_IN_Dataset} presents the evaluation results on the INTERACTION dataset. 
Similarly, the proposed model achieves the best performance on all four evaluation metrics. Its compact variant \emph{Ours (small)} remains competitive, exhibiting comparable \gls{fde} and outperforming all baselines in robustness. 

\textbf{\emph{Effects on Training Time}}:
As shown in Tables \ref{tab:model_comparison_DEEPSCENARIO} and \ref{tab:model_comparison_INTERACTION_IN_Dataset}, training time scales with model size and architectural complexity. Despite having more parameters, \emph{Ours} outperforms both \emph{GraphAIRL} baselines in terms of training times. \emph{Ours (small)} consistently trains faster than all agent-centric baselines, reducing training time by up to \SI{64.71}{\%} on DeepScenario and \SI{72.98}{\%} on INTERACTION without compromising performance. Moreover, while agent-centric models typically require more computational power, our compact model variant achieves training times on a T4 \SI{16}{\giga\byte} GPU comparable to the baselines trained on an A100 \SI{80}{\giga\byte} GPU.

\begin{figure}[t]
  \centering
  \vspace{-3pt}
  \includegraphics[width=0.44\textwidth]{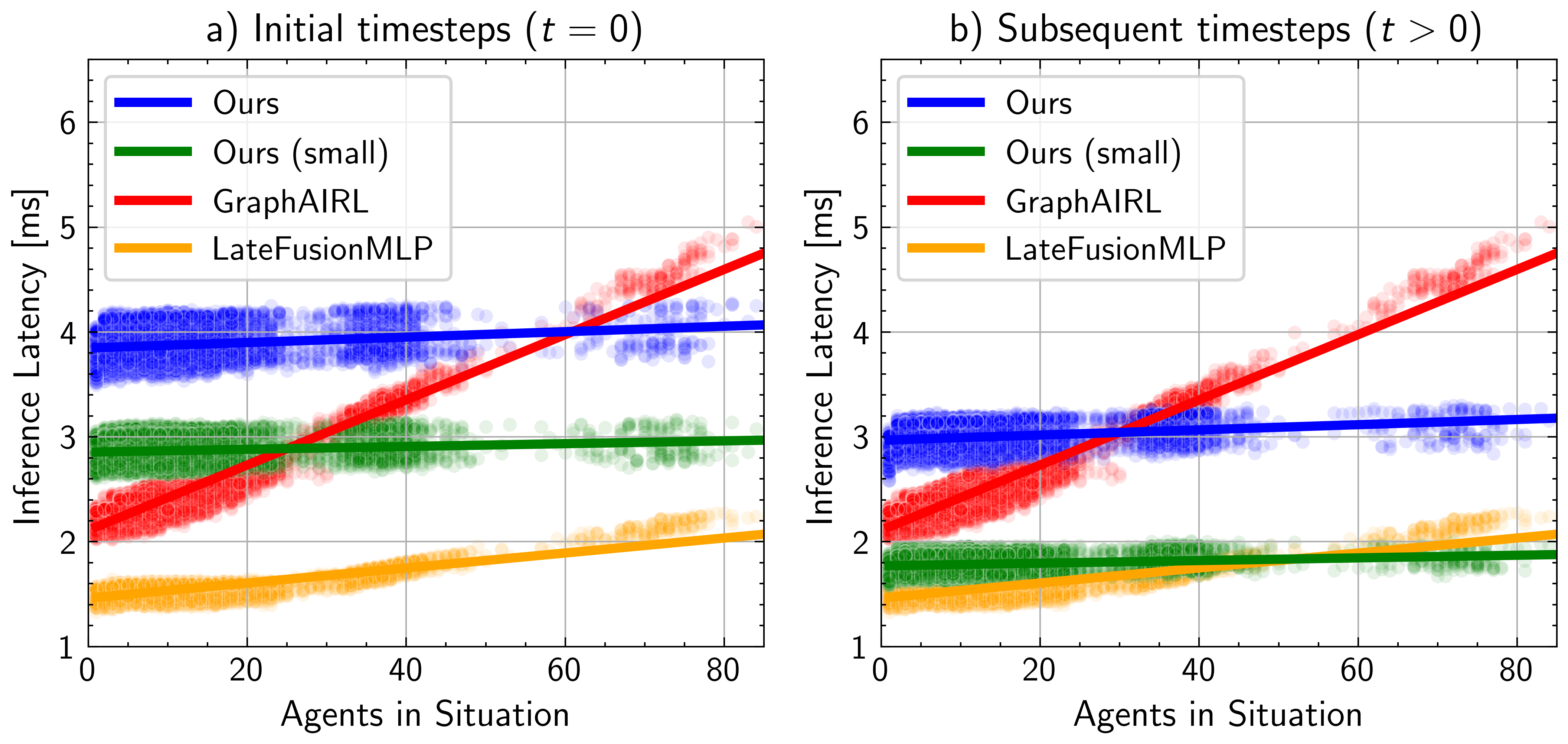}
  \vspace{-10pt}
  \caption{Regressed inference latency of a single policy-network forward pass \wrt the number of agents. a) initial simulation step. b) subsequent simulation steps. Measured on an A100 GPU. Dots denote individual scenes.}
  \label{fig:inference_times} 
  \vspace{-17pt}
\end{figure}

\textbf{\emph{Effects on Inference Time}}: 
Figure \ref{fig:inference_times} presents a quantitative comparison of single-step inference times across our models and baseline approaches. For the agent-centric models, inference time scales linearly with the number of agents, as observations must be encoded separately from each agent’s point of view. Nevertheless, the \emph{LateFusionMLP} consistently exhibits low inference latency due to its compact \gls{mlp}-based architecture. 
In contrast, our instance-centric approach decomposes inference into two stages: an initial simulation step, in which the entire map and all agents are encoded, and subsequent steps, in which the encoded map tokens are reused, substantially reducing the inference time for later simulation steps. As a result, simulating $n$ steps incurs the latency of the initial step only once, followed by the latency of subsequent steps $n-1$ times. For scenarios with few agents, the initial forward pass of our proposed small model is slightly slower than that of \emph{GraphAIRL}, since it encodes the entire map rather than only the regions visible to each agent. However, this overhead quickly diminishes as the number of simulation steps increases, yielding greater overall efficiency for longer simulations. Additionally, due to the symmetric scene context encoding, inference times in our instance-centric models stay nearly constant, regardless of the number of simulated agents.

\begin{figure}[t]
  \centering
  \vspace{2pt}
  \includegraphics[width=0.48\textwidth]{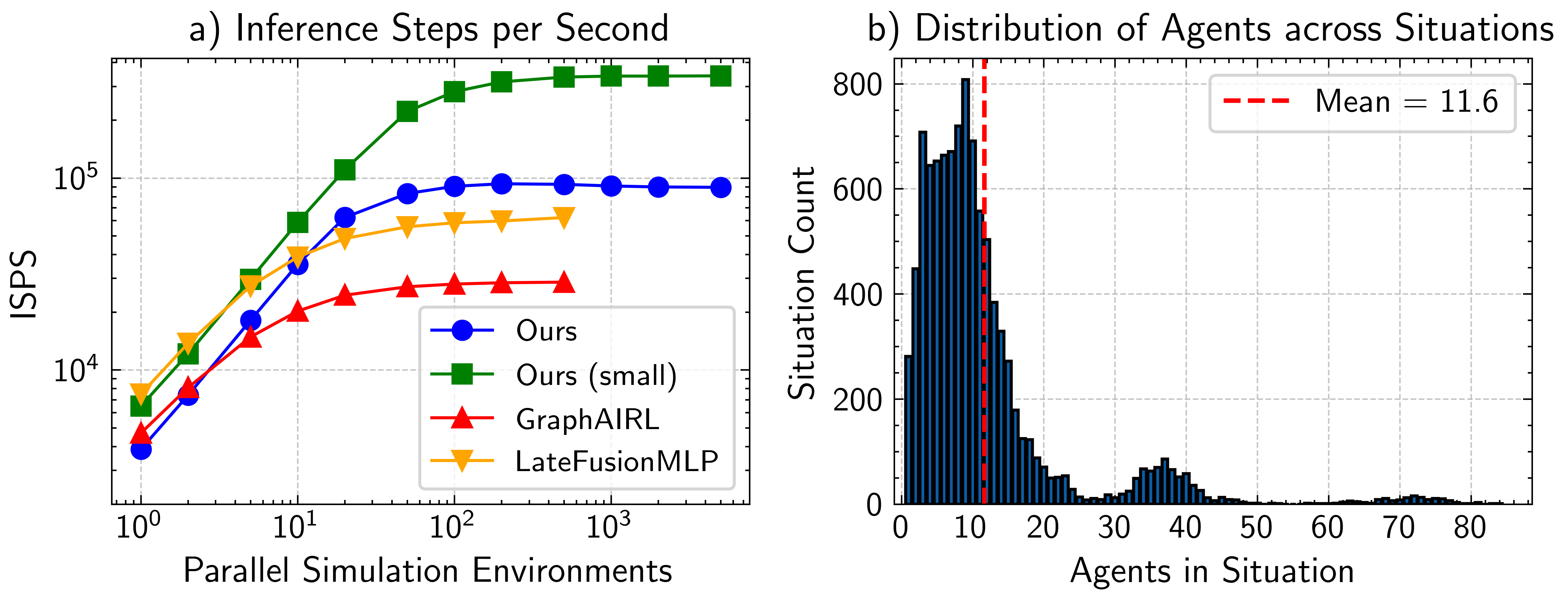}
  \vspace{-17pt}
  \caption{Peak throughput of the behavior model. a) Inference Steps per Second (ISPS) \wrt number of parallel simulation environments. b) Distribution of simulated agents across \num{10000} randomly sampled scenarios from the INTERACTION dataset. Measured on an A100 GPU.}
  \label{fig:isps}
  \vspace{-17pt}
\end{figure}

Commonly, multiple simulations are run in parallel. To evaluate the sample throughput of the behavior model under parallel simulation, we introduce the metric \emph{\gls{isps}}. This metric quantifies the number of agents across all simulation environments that can be processed by the behavior model per second. Formally, it is defined as $\mathrm{ISPS} = H \times \sum_{k=1}^N N_{A}^{(k)} / \Delta T$, where $N_{A}^{(k)}$ denotes the number of agents in the $k$-th simulation environment, $H$ is the number of inference steps taken, and $\Delta T$ is the elapsed time in seconds. Figure \ref{fig:isps} illustrates how the behavior model throughput scales with increasing numbers of parallel simulation environments as well as the distribution of simulated agents across $10^4$ randomly sampled scenarios. 
At smaller scales, all models benefit from parallelism, showing substantial gains in \gls{isps}. Agent-centric models saturate early at around \num{28.7}k and \num{60.3}k \gls{isps}, and cannot be scaled beyond \num{500} environments due to memory limits. In contrast, instance-centric models achieve higher throughout: \emph{Ours (small)} and \emph{Ours} exceed both agent-centric models with only five and twenty parallel environments, respectively, and continue scaling up to \num{5000} parallel environments. They reach peak throughput of about \num{93.4}k and \num{340.7}k \gls{isps}, representing an up to \num{11.9}-fold improvement over the agent-centric baselines.

\textbf{\emph{Effects on Cross-Dataset Performance}}: 
Next, we evaluate the models' performance in unseen traffic scenarios. For that, the models are trained on the DeepScenario dataset and subsequently evaluated on the INTERACTION dataset. As depicted in Table \ref{tab:model_comparison_INTERACTION_CROSS_Dataset}, all models trained on the DeepScenario dataset, which predominantly contains scenarios from Germany, experience notable performance drops when evaluated on the INTERACTION dataset, which includes diverse scenarios from Germany, China, and the USA. This effect is particularly pronounced in the baseline models. In contrast, our proposed instance-centric model demonstrates greater robustness, maintaining relatively low collision and off-track rates while exhibiting smaller increases in \gls{fde} compared to the baselines. We attribute this improved generalization to the viewpoint-invariant encoding of instances, which produces more transferable scene representations. By comparison, agent-centric approaches excel at modeling an individual agent’s local context but remain brittle to changes in viewpoint or environmental conditions, limiting applicability to unseen scenarios. Notably, the collision rate of our models increases only marginally, highlighting their capacity to better generalize to diverse interactions with other agents.

\begin{figure}[t]
  \centering
  \vspace{2pt}
  \includegraphics[width=0.48\textwidth]{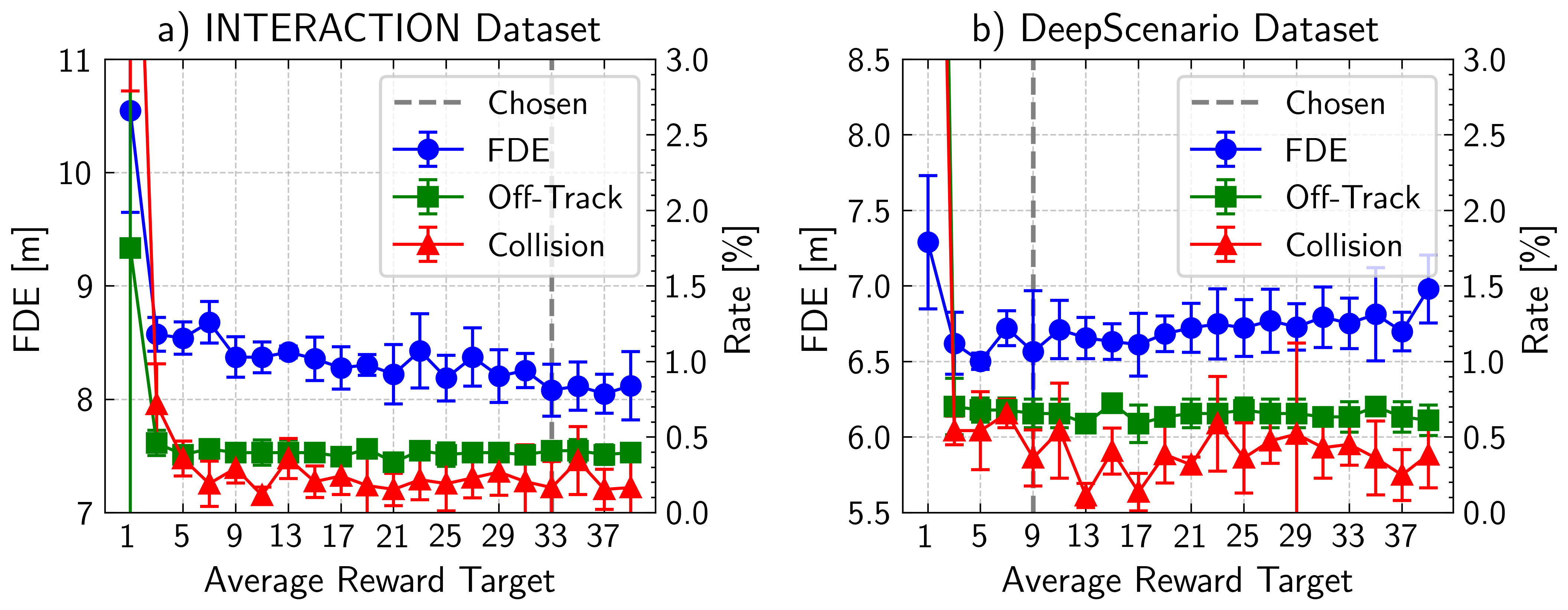}
  \vspace{-17pt}
  \caption{Ablation on the validation set \wrt the average reward target defined in \eqref{eq:target_mean} for a) INTERACTION dataset and b) DeepScenario dataset. Collision and off-track rates are calculated on the non-VRU subset. Results averaged over $5$ seeds. Whiskers indicate $\pm 1$ standard deviation.}
  \label{fig:ablation_tm}
  \vspace{-17pt}
\end{figure}

\textbf{\emph{Effects of Average Reward Target}}: 
Figure \ref{fig:ablation_tm} presents the results of our ablation study on different choices for the average reward target $\tilde{r}_\mathrm{target}$, evaluated across values from $1$ to $39$ in increments of $2$. In general, increasing $\tilde{r}_\mathrm{target}$ leads to more robust behavior, with larger gains being achieved at lower values. Initially, the positional accuracy, measured by \gls{fde}, increases significantly, demonstrating that the model benefits from the improved robustness. However, increasing $\tilde{r}_\mathrm{target}$ too much, leads to the positional error to rise again. For instance, this turning point can easily be seen for $\tilde{r}_\mathrm{target}=9$ on the DeepScenario dataset. This indicates a shift in the model's behavior, where it starts to prioritize robustness over realism. 
Based on these results, we selected the values $9$ for DeepScenario and $33$ for INTERACTION, representing the optimal trade-off between positional accuracy and robustness.

\textbf{\emph{Example Scenario}}: 
Figure \ref{fig:example_sit} shows a highly interactive example scenario featuring an all-way-stop intersection. All agents are controlled by our learned behavior model. Notably, the simulated and corresponding ground truth trajectories closely match, even over long horizons. The policy captures realistic driving dynamics, including similar speeds, accelerations, and subtle behaviors such as cooperative merging.

\section{Conclusion}
This work presents a novel approach to robust and efficient \gls{rl}-based behavior modeling for multi-agent driving simulation. Our instance-centric representation captures agents and map elements in their respective local coordinate frames, enabling viewpoint-invariant scene encoding. A Transformer-based refinement module captures interactions using relative positional encodings between reference frames. Static map elements need to be encoded only once per simulation, reducing redundant computation. As a result, both training and inference times are substantially reduced, and our model scales efficiently with the number of agents, making it particularly suitable for large-scale simulations. We obtain realistic behavior models via \gls{airl} and propose an adaptive reward transformation that decouples the agent's survival incentive from the discriminator performance, allowing fair comparisons across experiments. Across two automated driving datasets, our approach outperforms diverse baseline models in positional accuracy and robustness, and demonstrates significantly more generalizable behavior in cross-dataset evaluations.

\addtolength{\textheight}{-0cm}   




\section*{ACKNOWLEDGMENT}
The authors would like to thank Matthias Dingwerth and Matthias Steiner for their support and contributions to the development of the simulation framework.


\bibliographystyle{IEEEtran} 
\bibliography{library.bib}

\end{document}